\documentclass{article}

\usepackage{arxiv}

\usepackage[utf8]{inputenc} 
\usepackage[T1]{fontenc}    
\usepackage{hyperref}       
\usepackage{url}            
\usepackage{booktabs}       
\usepackage{amsfonts}       
\usepackage{nicefrac}       
\usepackage{microtype}      
\usepackage{lipsum}
\usepackage{authblk}

\usepackage{float} 
\usepackage{times}
\usepackage{epsfig}
\usepackage{graphicx}
\usepackage{amsmath, amsfonts, amssymb}
\usepackage{bm}
\usepackage{algorithm}
\usepackage{algorithmicx}
\usepackage{comment}
\usepackage{fullpage}
\usepackage{tikz}
\usepackage{pifont}
\usepackage{amsthm}
\usepackage{subcaption}
\newtheorem{definition}{Definition}

\usepackage{amsmath,amsfonts,amssymb}
\usepackage{mathtools}
\usepackage{multirow}
\usepackage[noend]{algpseudocode}
\usepackage{booktabs}
\usepackage[export]{adjustbox}
\usepackage{url}

\providecommand{\keywords}[1]{\textbf{\textit{Index terms---}} #1}

\newcommand{\etal}{\textit{et al}.~}
\usepackage{xspace}
\newcommand*{\eg}{\textit{e.g.}\@\xspace}
\newcommand*{\ie}{\textit{i.e.}\@\xspace}

\newcommand*{\vs}{\textit{vs.}\@\xspace}
\newcommand*{\wrt}{\textit{w.r.t.}\@\xspace}
\makeatletter
\newcommand*{\etc}{%
	\@ifnextchar{.}%
	{\textit{etc}}%
	{\textit{etc.}\@\xspace}%
}
\makeatother
\algtext*{EndWhile}
\algtext*{EndFor}
\algtext*{EndIf}
\algdef{SE}[DOWHILE]{Do}{doWhile}{\algorithmicdo}[1]{\algorithmicwhile\ #1}%

\makeatletter
\def\BState{\State\hskip-\ALG@thistlm}
\makeatother

\title{TIME: A Transparent, Interpretable, Model-Adaptive and Explainable Neural Network for Dynamic Physical Processes}

\author[1]{\textbf{Gurpreet Singh} \textsuperscript{\dag}}
\author[2]{\textbf{Soumyajit Gupta} \textsuperscript{\dag}}
\author[3]{\textbf{Matt Lease}
\textsuperscript{\ddag}}
\author[1]{\textbf{Clint N. Dawson} \textsuperscript{\ddag}}
\affil[1]{Oden Institute for Computational Engineering and Sciences}
\affil[2]{Department of Computer Science}
\affil[3]{School of Information}
\affil[ ]{University of Texas, Austin}
\affil[ ]{\texttt{\{gurpreet, smjtgupta, ml\}@utexas.edu, clint@oden.utexas.edu}}

\begin{document}

\maketitle

{\let\thefootnote\relax\footnote{{\dag contributed equally to this work under the supervision of \ddag.}}}

\begin{figure}[h]
	\centering
	\includegraphics[width=0.95\linewidth]{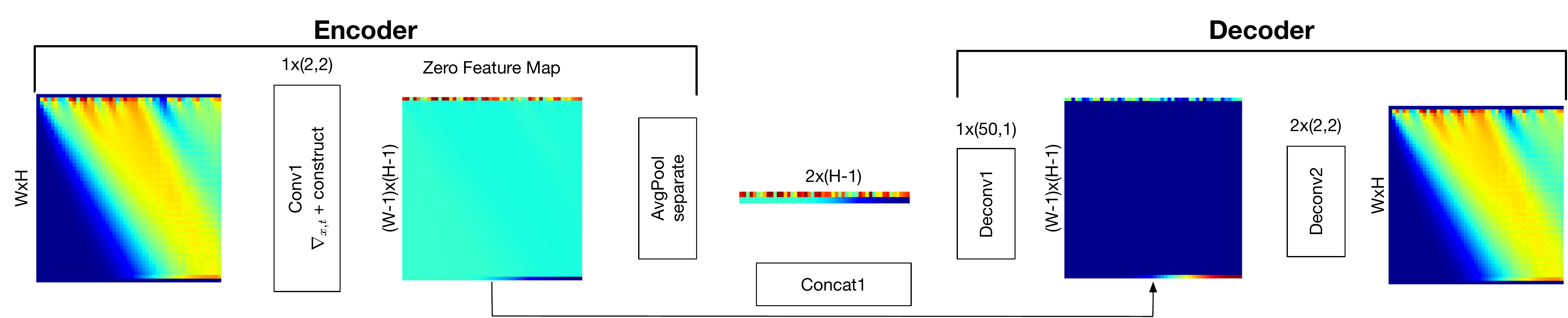}
	\caption{A single layer encoder-decoder for a two-dimensional hyperbolic problem with complete observation data.}
	\label{fig:1d_hyperbolic}
\end{figure}
\begin{abstract}
	Partial Differential Equations are infinite dimensional encoded representations of physical processes. However, imbibing multiple observation data towards a coupled representation presents significant challenges. We present a fully convolutional architecture that captures the invariant structure of the domain to reconstruct the observable system. The proposed architecture is significantly low-weight compared to other networks for such problems. Our intent is to learn coupled dynamic processes interpreted as deviations from true kernels representing isolated processes for model-adaptivity. Experimental analysis shows that our architecture is robust and transparent in capturing process kernels and system anomalies. We also show that high weights representation is not only redundant but also impacts network interpretability. Our design is guided by domain knowledge, with isolated process representations serving as ground truths for verification. These allow us to identify redundant kernels and their manifestations in activation maps to guide better designs that are both interpretable and explainable unlike traditional deep-nets.
\end{abstract}

\section{Introduction}
Convolutional neural networks \cite{krizhevsky2012imagenet} have become an indispensable tool for a wide range of applications due to their flexibility in learning good feature representations. With the pervasive use of CNNs in all aspects of our lives, model interpretability is a crucial issue besides model accuracy \cite{bau2017network}, when we need people to trust a network's prediction. In spite of high accuracy of neural networks, human beings cannot fully trust a network, unless it can explain its logic for decisions \cite{zhang2018interpretable}. However, interpretability comes at the cost of accuracy \cite{lakkaraju2016interpretable}.
	
Physical sciences determine an invariant structure in observations and its subsequent parametrization as models (condensed representations) to reproduce them. This invariant structure is \wrt the description of a space-time continuum hypothesis where the observations are in turn embedded. Assuming physical locality this invariant structure can be extracted with limited observations from human-isolated physical processes resulting in decoupled models. The advent of classical compute machines and consequent numerical solution algorithms expanded the scope of our understanding with coupled PDEs as condensed representations of multiple physical processes occurring simultaneously in a space-time domain. These PDEs were the earliest representations of the invariant structure/adjacency map that allowed us to consequently compress these space-time observations under the assumption that they are indeed correlated. 

\begin{figure}[h]
	\centering
	\includegraphics[width=0.9\linewidth]{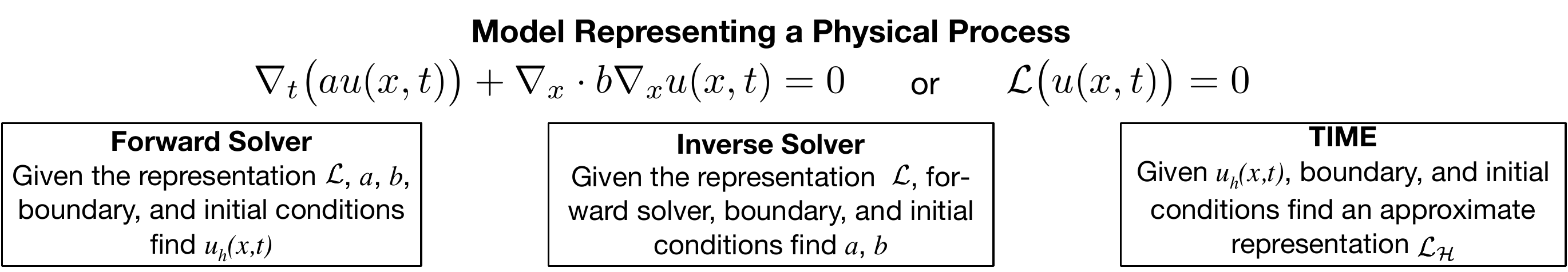}
	\caption{An overview of PDE solvers and what we are trying to achieve with our architecture. Here, $\nabla_{t}$, $\nabla \cdot$, $\nabla_{x}$ carry their usual meaning as time differentiation, divergence and gradient operators respectively. $u(x,t)$ is the space-time observable and $a,b$ are coefficients.}
	\label{fig:time}
\end{figure}
	
We explicitly state that our intent is not to develop a computationally efficient PDE solver. We are only interested in identifying encoded representations with discrete kernels approximating the operators employed in PDEs. An extensive amount of literature is available on finite element, finite difference, and finite volume schemes \cite{babuska2012modeling,thomee2001finite} that accurately and efficiently solve discrete forms of known PDEs. Our focus in this proposed research is to develop encoders with the decoder serving as a visual tool to ratify the encoded learning. Fig. \ref{fig:time} gives a visual representation of our interpretation of modeling in physical sciences and the data-guided evolution of understanding proposed in this work. 

A number of convolutional architectures are available in the literature that attempt to represent dynamic physical processes. However, these networks assume no distinction between spatio-temporal operators. This leads to a loss in human interpretability of the learned representations. For \eg ODE-net \cite{chen2018neural} identifies a first order temporal operator to relate Resnet architecture to the description of an Ordinary Differential Equation (ODE). The spatial distribution is then learned as a single neural network block making the learned kernels difficult to decipher or independently verify. On the other hand, PDE-net \cite{long2017pde,long2019pde} attempts to learn the spatial operators separately. However, one must note that the numerical experiments do not clearly show how the learned kernels of size (3x3, 5x5) represent these linear operators in space.
	
\begin{figure}[ht]
	\centering
	\begin{subfigure}[b]{0.54\linewidth}
		\centering
		\includegraphics[width=0.7\linewidth]{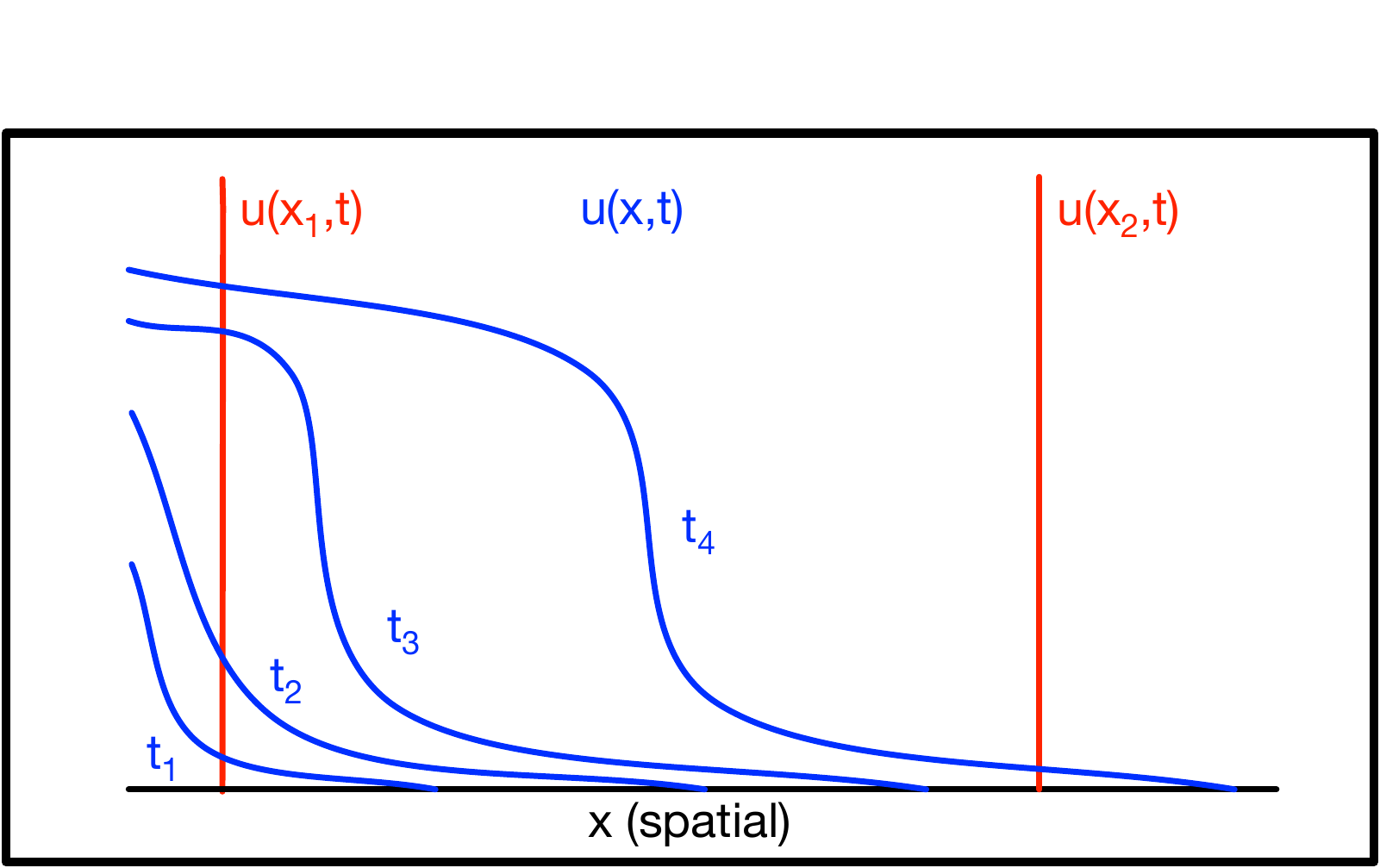}
	\end{subfigure}
	\begin{subfigure}[b]{0.45\linewidth}
		\centering
		\includegraphics[width=0.48\linewidth]{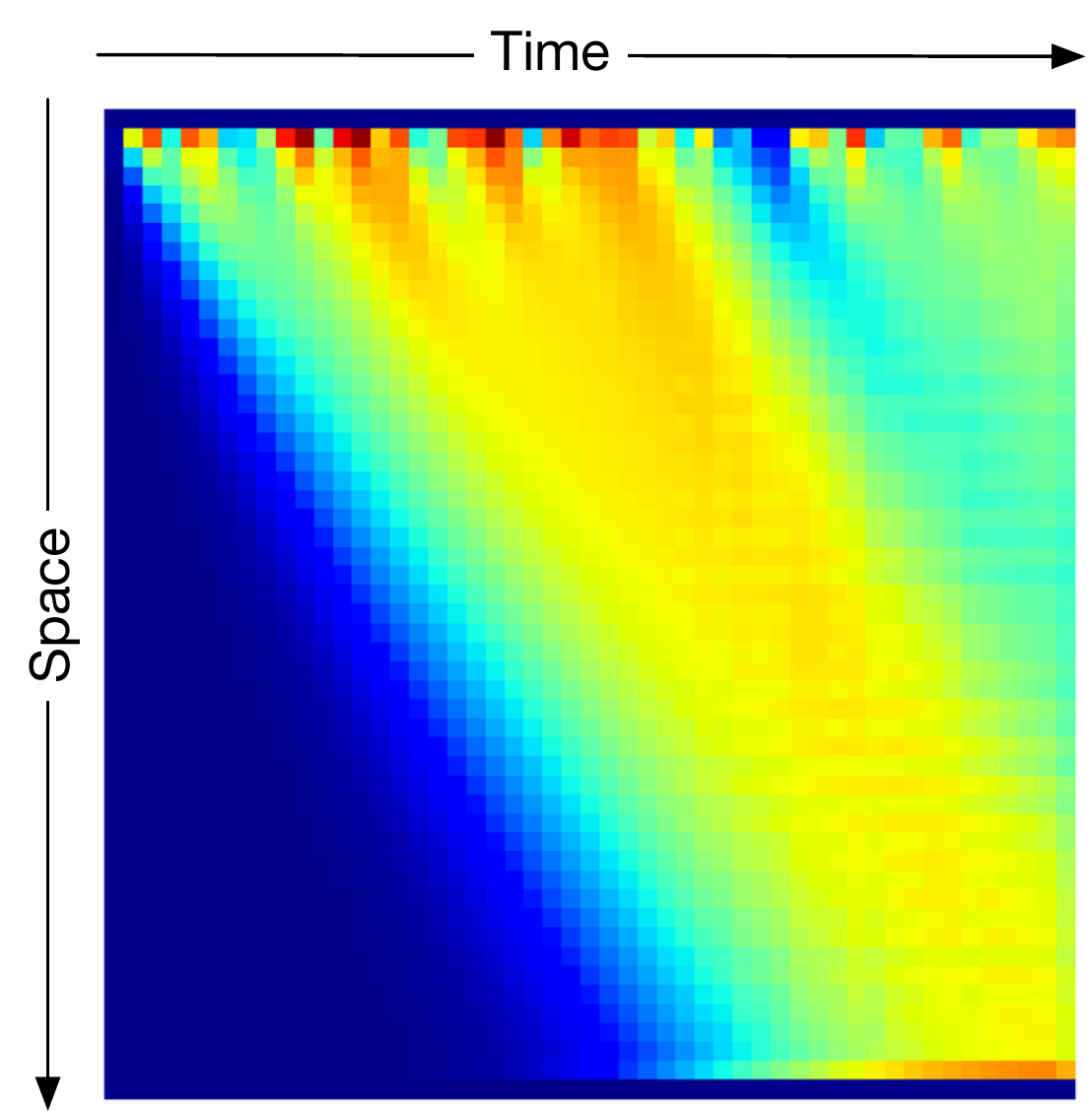}
	\end{subfigure}
	\caption{Left: Temporal iso-maps of an observation $u(x,t)$ for a first order hyperbolic shock traversal. Right: Corresponding space-time image generated using random perturbations propagating as waves with appropriate padding.}
	\label{fig:overview}
\end{figure}
	
In this work, we present a low-weight architecture that takes into account the aforementioned considerations learnt from our fellow researchers. Our fully convolutional architecture learns higher order operators as the depth of the network increases. We exploit convolutional layers as control points for extracting interpretable representations. Extracting non-linearities as functions of state variables using convolutional layers as control points will be a topic of future research. Dense networks are also capable of learning similar representations however a lack of interpretation of the learned kernels obfuscates detailed verification and validation. The main contributions of the paper are as follows:
\begin{enumerate}
	\item Sparse Kernel Embedding: A non-causal, fully convolutional network architecture with encoding kernels.
	\item Model Interpretability: Encoder kernels approximating the physical process.
	\item Model Explainability: Boundary conditions introduces scaling in kernels as external perturbation, and detect anomalies due to missing or deviatory data.
	\item Model Transparency: Network depth (\#layers) and (\#feature maps) in each layer are guided by the order of physical processes we are trying to capture. 
	\item Model Adaptive: Can resolve both isolated and coupled processes.
	\item Fully Convolutional end-end: Independent of the domain size
\end{enumerate}
	
\section{Problem Statement}

The primary objective of this work is to develop an encoder-decoder that identifies the invariant structure as an adjacency map of a dynamic physical process. The research goal is to learn improved representations (models) by imbibing sensor data from multiple sources over a space-time domain ($\Omega \times (0,T]$) under observation. We rely upon the hypothesis; later demonstrated using numerical experiments, that a representation (or model) satisfies the following identity:

\begin{equation}
\mathcal{L_H}(\mathbf{u}_{h}) = 0.
\label{eqn:identity}
\end{equation}
	
Here, $h$ indicates finite dimensional observation data and $\mathcal{H}$ the order of the approximating kernel. Further, $\mathbf{u}_{h}$ is the vector of all observations and $\mathcal{L_H}$ is a discrete functional operator on the observation vector densely encoding the underlying spatio-temporal invariant structure. We assume that: (1) $\mathcal{L_H}$ is a spatio-temporal operator to be purely inferred from the observation data, and (2) the boundary and initial conditions, and forcing functions are a subset of the observation tensor $\mathbf{u}_{h}$. The former assumption avoids explicit specification of the directional nature of the feature being observed. This, in turn allows us to infer locality in the spatio-temporal domain, to develop a general framework that can represent a larger class of physical processes. The latter consideration results in an unsupervised learning framework where the input ($\mathbf{u}^{i}_{h}$) and output feature ($\mathbf{u}^{o}_{h}$) vectors are such that $\mathbf{u}^{i}_{h}\cup\mathbf{u}^{o}_{h}=\mathbf{u}_{h}$. We now pose the following questions in order to imbibe physical arguments and practical considerations leading to the construction of a CNN framework:

\begin{enumerate}
	\item {Can we recover an invariant structure given dense space-time observations?}
	\item {Is it possible to detect the nature of missing observations or anomalies?}
	\item {How can we make the network more interpretable and transparent?}
	\item {How can we evolve a representation without losing interpretability?}
\end{enumerate}
	
We first attempt to answer each of these questions separately by considering simple examples towards weaving a unified CNN framework called TIME. Our core objective is to develop  domain-informed interpretability to bridge the gap between between practitioners (experimentalists) and developers (modelers). Please note that in the following by kernels ($L$) we mean the convolutional kernels to separate it from the definition of mathematical kernel in linear algebra $\mathcal{L}(u) = 0$ that maps an input feature to a null space.  

Let us consider the simple encoder-decoder network in Fig. \ref{fig:1d_hyperbolic} for a first-order linear hyperbolic problem. The training data is chosen to be a space-time image (second order tensor $(W \times H)$) of the characteristic with training labels (second order tensor $(2 \times H)$) prescribed by the point-wise source terms at $x=0$ and $x=L$ spatial boundaries. Our objective is to find a representation such that $\mathcal{L_H}(u_{h})=0$ where $u_{h}$ is the concatenated observation data. This simplistic problem at hand only requires a single convolutional layer, encoder with 4 parameters in Eq. \ref{ker:hyp} to learn the discrete operator $\mathcal{L_H}$. 
Furthermore, the observation data can be decomposed into different combinations of training image and labels stemming from practical considerations as long as $\mathbf{u}^{i}_{h}\cup\mathbf{u}^{o}_{h}=\mathbf{u}_{h}$.
	
The key purpose of this demonstration is to elucidate a \textit{zero-feature map} that allows us to check if the learned representation satisfies the identity in Eq. \ref{eqn:identity}. Although the exact kernel (without scaling) in Eq. \ref{ker:hyp} can be computed trivially for this problem and used for verification, the \textit{zero-feature map} will be later used to learn correct representations wherein the exact kernels composing the representation (or model), are either not known or difficult to compute a priori.
\begin{equation}
ker = \begin{bmatrix}
+0 & -1\\
-1 &  +2
\end{bmatrix},
\label{ker:hyp}
\end{equation}

In the following, we present a fully convolutional architecture where the encoding kernels compose a discrete PDE operator $\mathcal{L_H}$ given space-time observation data. The decoding, transpose convolutional, kernels then represent the interpolation operators that approximately reconstructs input observation data $u^{i}_{h}$ from the learned encoding kernels. We consider synthetic datasets generated using a forward PDE solver for four benchmark problems, described below, to verify our proposed architecture. In the following $a$ and $b$ are known constants with $f$, $g$, and $p$, $q$ as known functions in space and space-time, respectively. 
	
\subsection{Hyperbolic PDE}
The hyperbolic PDE used to generate the data set on a spatio-temporal domain $\Omega \times (0,T]$ is given by:
\begin{align}
\nabla_{t} \big(b v(x,t)\big) - \nabla_{x} \cdot a v(x,t) = q(x,t)  \quad \text{in} \, \Omega \times (0,T]
\label{eqn:hyp}
\end{align}
Here, subscripts $\{x,t\}$ denote the spatial or temporal operator, with $\nabla$ and $\nabla \cdot$, the gradient and divergence (directional gradient), respectively. Further, $a$ and $b$ are known constants with $q(x,t)$ a prescribed function as a point-wise perturbation. $v(x,t)$ is the characteristic that propagates in space with a known velocity $a$. The resulting solution is a space-time image shown in Fig. \ref{fig:overview}. For the sake of completeness, the boundary (spatial) and initial (temporal) conditions on $v(x,t)$ are given by:
\begin{align*}
\nabla_{x} \big( a v(x,t) \big) \cdot n = 0  \quad \text{on} \, \partial \Omega \times (0,T]\\
v(x,0) = g(x,0) \quad \text{on} \,  \Omega \times \{0\}
\end{align*}
Here $n$ is a prescribed unit normal in the spatial direction. Random perturbations are introduced through boundary condition specification at $x=0$ with an open boundary at $x = L$. In the discrete form, the same can be achieved for this system by imposing point-wise perturbation $q(x,t)$ at the two spatial boundaries.
	
\subsection{Elliptic PDE}
The elliptic, Poisson or diffusion equation is given by: 
\begin{align}
\nabla_{x} \cdot a \nabla_{x} u(x) = p(x) \quad \text{in} \, \Omega
\label{eqn:elp}
\end{align}
The invariant structure here is sparse due to the absence of temporal operators making it a good verification problem for our numerical experiments. Since the solution is time invariant only spatial boundary condition is prescribed as follows:
\begin{align*}
\nabla_{x} u \cdot n = 0  \quad \text{on} \, \partial \Omega\\
\int p(x) = 0  \quad \text{on} \, \Omega
\end{align*}
The last equation represents a compatibility condition to ensure the solution remains unique. The absence of this condition implies the solution $u(x,t)$ will only be unique up to a constant.

\subsection{Parabolic PDE}

The parabolic PDE then allows us to get a spatially and temporally varying solution for $u(x,t)$ given by:
\begin{align}
\nabla_{t} \big(b u(x,t)\big) - \nabla_{x} \cdot a \nabla_{x} u(x,t) = p(x,t)  \quad \text{in} \, \Omega \times (0,T]
\end{align}
under the spatial and temporal boundary conditions as:
\begin{align*}
\nabla_{x} u(x,t) \cdot n = 0  \quad \text{on} \, \partial \Omega \times (0,T]\\
u(x,0) = f(x,0) \quad \text{on} \,  \Omega \times \{0\}
\end{align*}
	
\subsection{Coupled elliptic hyperbolic PDE}

The coupled elliptic and hyperbolic problem lets us expand our learning and network architecture. Here, both elliptic and hyperbolic kernels must be learned in order to represent the underlying invariant structure.  The synthetic dataset for the coupled system is generated using the following system of partial differential equations:
\begin{align}
\nabla_{x} \cdot a \nabla_{x} u(x,t) = p(x,t) \quad \text{in} \, \Omega \times (0,T] \\
\nabla_{t} \big(b v(x,t)\big) - \nabla_{x} \cdot \big( u(x,t) v(x,t)\big) = q(x,t)  \quad \text{in} \, \Omega \times (0,T]
\end{align}
The boundary and initial conditions used to generate the dataset are given by:
\begin{align*}
\nabla_{x}u(x,t) \cdot n = 0 \quad \text{on} \, \partial \Omega \times (0,T]\\
\nabla_{x} \big( u(x,t) v(x,t) \big) \cdot n = 0  \quad \text{on} \, \partial \Omega \times (0,T]\\
u(x,0) = f(x,0) \quad \text{on} \,  \Omega \times \{0\}\\
v(x,0) = g(x,0) \quad \text{on} \,  \Omega \times \{0\}
\end{align*}
This coupled problem allows us to expand our network architecture guided by domain knowledge and application oriented considerations later discussed in the results section.
	
\section{Related Work}

Early methods of visual explanation involved salience map approach which highlighted important pixels. Zeiler \etal \cite{zeiler2014visualizing} is based on occlusion, where a network is tested with occluded inputs to create a map showing which parts of the data that influences the output. Simoyan \etal \cite{simonyan2014very} produces the map by directly computing the input gradient through back propagation. Other approaches include Grad-CAM \cite{selvaraju2017grad}, SmoothGrad \cite{smilkov2017smoothgrad} \etc. describing both high activation (where neurons fire strongest) and high sensitivity (where changes would most affect the output) areas \cite{gilpin2018explaining}. Zhang \etal \cite{zhang2018interpretable} enforces explanation in the activation maps through a mutual information criteria. Traditionally, interpretation is considered to be inversely related to model accuracy. However, we note that a loss in accuracy can also be due to the choice of a regularizer that we impose on the network. These choice stem from a black-box outlook of the network design (depth, width, and kernel size) where beliefs are latter enforced on the activation maps as regularizers to enforce human understanding. In our work, we rely upon the kernels for interpretations and activation maps for explanations to guide design choices. 
	
\section{Framework}
Although interpretation, explanation and transparency are considered as highly correlated by Lipton \cite{lipton2016mythos}, we propose our own definition \wrt domain knowledge inherent to the physical process. We do not treat our network as a black box, rather, it is open to inspection by a modeler so as to examine and validate the kernels, activations and scales. Same goes for explainability which has been traditionally seen as post-hoc interpretations of the network. We adapt these definitions from literature but give a polished domain-guided meaning to these words to highlight how they correlate to our problem. To set a sense, consider a simplistic hyperbolic PDE in Eq. \ref{eqn:hyp}.

\begin{definition}
	An Interpretation is the mapping of an abstract concept (spatio-temporal operators) into a domain that a human can make sense of \cite{montavon2018methods}.
\end{definition}
	
The images itself give a very broad overview of the process and as later shown for coupled processes become increasingly tedious to directly interpret. However, a simple kernel correlating the concept of change in space and time in an observed quantity is easier to grasp. This is equivalent to identifying the $\nabla_{x,t}$ operator in Eq. \ref{eqn:hyp} from the gradient kernel Eq. \ref{ker:hyp} considering only the sign of the learned kernel weights without the scaling. This is especially true for physical processes where visualizing a phenomenon is often restrictive.

\begin{definition}
	An Explanation is a collection of features in the interpretable domain, that has contributed for a given example to produce a decision \cite{montavon2018methods}.
\end{definition}

These are equivalent to identifying the scaling resulting from coefficients $(a, b)$ in Eq. \ref{eqn:hyp} and the boundary conditions, reconstructed space time images in Fig. \ref{fig:1d_hyperbolic} which a human can validate as outputs of the system. The factors scaling ($a, b$) the kernels are the characteristics of a system under observation and their ranges can be validated from experimental data under isolated conditions. For example, in dynamic physical processes these usually represent a material property that might have a spatial distribution but remains temporally invariant. 

\begin{definition}
	Transparency is a sense of understanding of the mechanism by which the model works. It can be defined both at the level of the entire model (simulatability) and at the level of individual components (decomposability) \cite{lipton2016mythos}.
\end{definition}
	
For a model to satisfy \textit{Simulatability}, the modeler should be able to take the input data together with the parameters of the model to reproduce the boundary conditions in limited time. To satisfy \textit{Decomposability}, each part of the model (feature maps and weights) admits an intuitive explanation as discussed in Section \ref{sec:ker_rep}. The clear semantics in the convolution layers are of great importance when we need people to trust a network's prediction \cite{zhang2018interpretable}. Usually, when we try to make a CNN interpretable, that comes along with a loss in accuracy (its discrimination power). However, we show that since the layers of the proposed model are domain-guided, it achieves better predictive capability. We consider the isolated process kernels as an ideal state, a deviation from which allows us to identify the evolved representation. The learned kernels for the coupled process indicate shared characteristics with elliptic and hyperbolic problems. This renders our architecture transparent as deviation in learned kernels from their isolated process counterparts.
	
\section{End-End Encoder Decoder Network}

 A fully convolutional architecture with a zero-feature map allows us to identify a representation that conserves the observable quantities in the input feature map. For the benchmark problems at hand, this implies mass or energy conservation depending on the application domain: fluid flow, component transport, heat advection and diffusion \etc. Since the datasets are synthetically generated, we are able to verify this directly. However, for practical applications if the observation data is sparse or missing, the representation will be conservative up to the training/validation loss.
\subsection{Datasets}

Data is generated for four PDE configurations: hyperbolic, elliptic, parabolic, and coupled hyperbolic-elliptic processes. Each family has $101$ images of size $(W,H)=(50,50)$ in a $(space, time)$ format and has $2$ channels corresponding to $u(x,t)$ and $v(x,t)$. Each image also has boundary information at the injection and production point of size $(50 \times 2)$. The data is split in a $80\%/20\%$ for training/validation phase. We do not perform any data augmentation.

\subsection{Architecture}

\begin{figure}[th]
	\centering
	\includegraphics[width=0.9\linewidth]{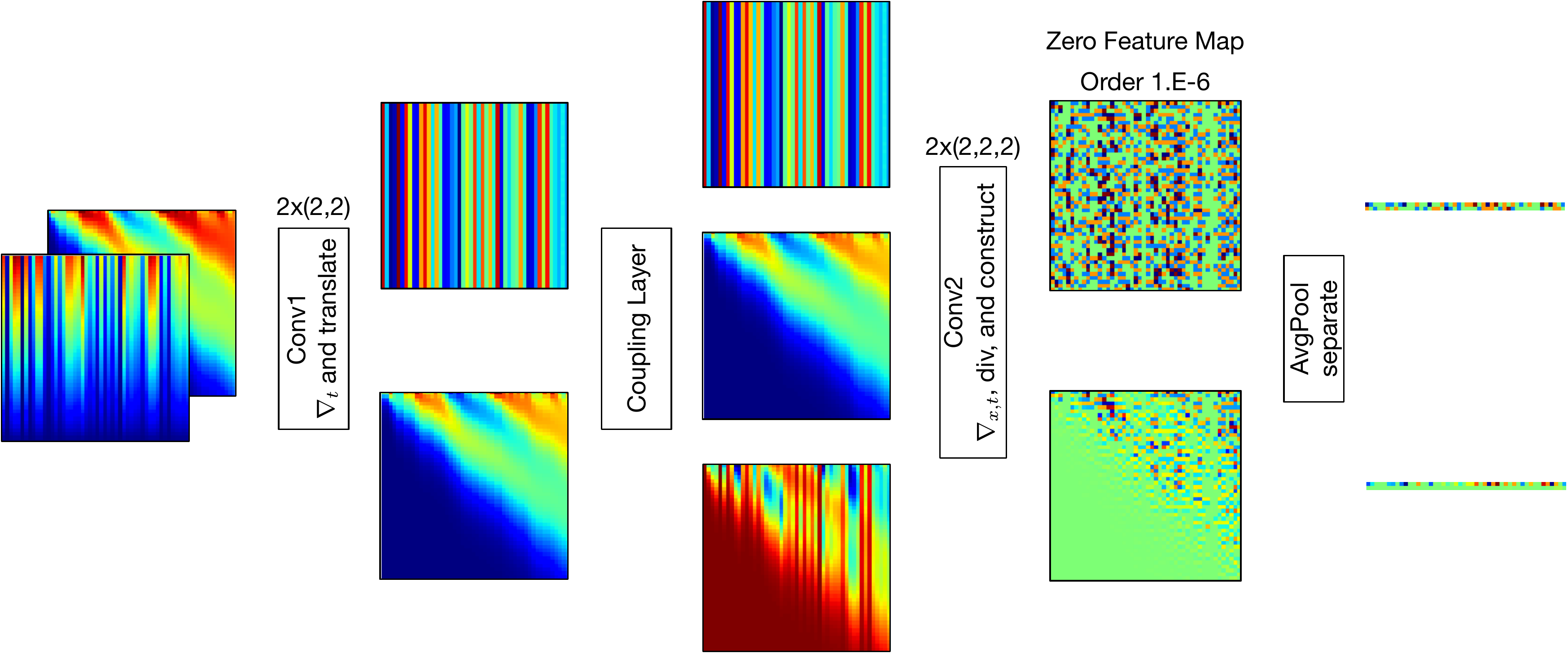}
	\caption{Encoder for a coupled process with elliptic and hyperbolic forms in the input channels. The coupling layer allows the elliptic and hyperbolic forms to interact. The network is capable of representing isolated elliptic and hyperbolic input maps as well at the cost of redundant kernel evaluations.}
	\label{fig:1d_coupled}
\end{figure}

The design of our end-to-end convolutional model is motivated by the infinite dimensional PDEs. It consists of encoder and decoder modules with skip connections between them as shown in Fig.~\ref{fig:1d_hyperbolic}, and the boundary layer, for isolated system. For coupled system in Fig. \ref{fig:1d_coupled}, an additional coupling layer is present.

The network takes in a (space-time) image $\mathbf{I}$ of size $(W,H)$. The image is passed through a stack of convolutional  layers with filters having a small receptive field: $\mathbf{2 \times 2}$, with a stride of $1$ pixel. Padding is set to `valid' to avoid zero padding at the boundaries. Average pooling is done at the encoder layer end on the spatial axis since it tries to capture the boundary layer representation of the flow itself, giving a $(2, H-2)$ pixel output. All hidden layers have `tanh' activation. There are no biases used in the network. A general mapping of the network is shown in Table \ref{tab:summary_params}. Coupling layer allows two different channels to interact to account for the physical coupling between processes. For Fig. \ref{fig:1d_coupled} the third channel is a multiplication of the other two channels to capture the interaction between elliptic and hyperbolic forms.

The depth of the network is guided by the order of the PDE, hence a $n^{th}$ order PDE will have $n$ \textit{conv} layers. The number of kernels in each layer is guided by the number of operators of a certain degree. Since the coupled system in Fig. \ref{fig:1d_coupled} is a second order system, the encoder depth is set to $2$, and kernels at each depth are set to $2$ for capturing gradients across two variables. The decoder module contains concatenated layers which arrive as skip connections from the encoder. This helps to stabilize to decoder since it tries to reconstruct the space-time from just the boundary conditions. 
\begin{table}[ht]
	\centering
	\caption{Overview of network architecture for isolated and coupled processes. $K_1, K_2$ are the number of kernels chosen for layers $1,2$, respectively}
	\begin{tabular}{|c|c|c|c|c|} \hline
	    Process & \multicolumn{2}{c|}{Isolated} & \multicolumn{2}{c|}{Coupled}\\ \hline
		Layer & Output Shape & \# Params & Output Shape & \# Params \\ \hline 
		Input & $(W,H,1)$ & $0$ & $(W,H,2)$ & $0$ \\
		Conv1 & $(W-1,H-1,K_1)$ & $4 K_1$ & $(W-1,H-1,K_1)$ & $8 K_1$ \\
		Coupling & - & - & $(W-1,H-1,K_1+1)$ & 0 \\
		Conv2 & $(W-2,H-2,K_2)$ & $4 K_1 K_2$ & $(W-2,H-2,K_2)$ & $6 K_1 K_2$ \\
		Encoder & $(2,H-2,1)$ & $0$ & $(2,H-2,2)$ & $0$ \\
    \hline
	\end{tabular}
	\label{tab:summary_params}
\end{table}
The choice of a fully convolutional network as opposed to other ConvNets with dense layers is four-fold: (1) The width of dense layers depend on the output feature map thereby making the network itself problem dependent. A fully convolutional choice lets us handle input feature maps of arbitrary size $(W \times H)$ without any loss of generality. (2) Dense layers after flattening the feature map results in a high weights representation which we would like to avoid. (3) This in turn results in high-dimensional minimization problem. Given the fact that our representations are sparse, this requires a regularizer or dropout at the cost of additional computations. (4) Usage of multiple dense layers after convolutional layer results in loss of interpretability.
	
\subsection{Setup and training}

All experiments were done on a setup with Nvidia 2060 RTX Super 8GB GPU, Intel Core i7-9700F 3.0GHz 8-core CPU and 16GB DDR4 memory. We use the Keras \cite{chollet2015} library running on top of a Tensorflow 1.14 backend with Python 3.5 to train the neural networks in this paper. The network is trained using the above setup and for optimization we use \textit{AdaDelta} \cite{zeiler2012adadelta} with parameters $(lr=0.85,rho=0.95,decay=0)$. Without a domain-guided initialization, training takes about $8$ min and $70$ epochs, with $2000$ steps in each epoch, to converge to the true solution. A single forward pass of the network takes 75 ms. A Mean-Squared Error (\textit{MSE}) loss is used for comparing the predicted boundary conditions against the ground truth labels. A zero-sum regularizer is imposed on the process kernels to replicate a physical process satisfying Eq. \ref{eqn:identity}.

\subsection{Results}

\begin{figure}[h]
    \centering
		\includegraphics[width=0.9\linewidth]{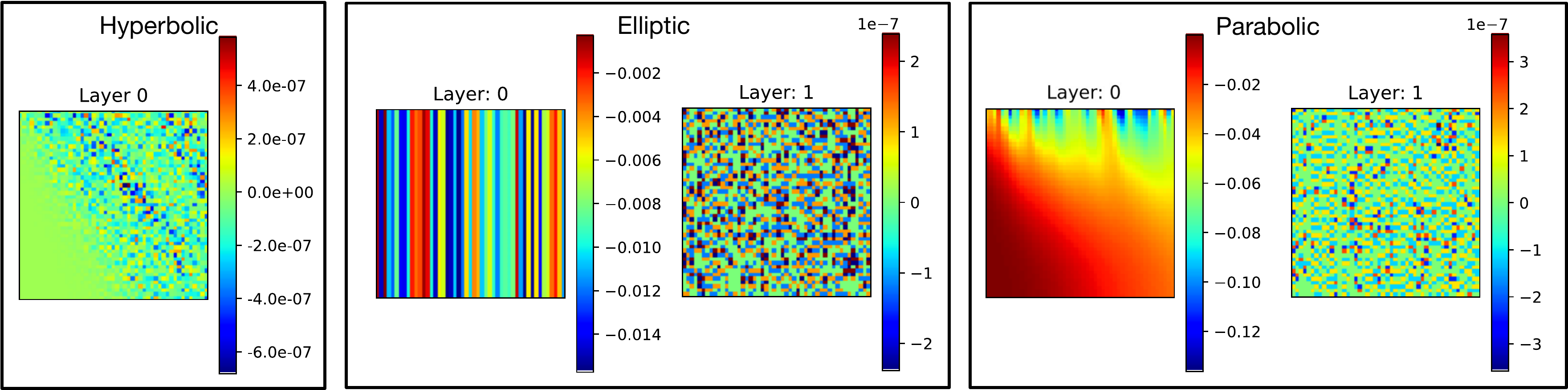}
    \caption{Activation maps for the three isolated processes.}
    \label{fig:act_isolated}
\end{figure}

Evaluation is done on the four families of PDEs. Fig. \ref{fig:loss} shows the high rate of convergence due to proper design of the network layers. MSE reached $\sim e-11$ around 60 iterations and it was ran till $150$ iterations where it stabilized at $\sim e-12$. The ground truth MSE with true kernels is $\sim e-14$. Activation maps for different isolated PDEs are shown at scale in Fig. \ref{fig:act_isolated}, where they reach $\sim e-7$ at convergence. The low-weight property of the network can be seen in Table \ref{table:params}.

\begin{table}
    \begin{minipage}{0.45\linewidth}
		\caption{Network parameters for the encoder corresponding to Table \ref{tab:summary_params}}
		\label{table:params}
		\centering
		\begin{tabular}{|c|c|c|} \hline
			Data Type & I/p Shape & \# Param \\ \hline 
			Hyperbolic & $(W,H,1)$ & $8$ \\
			Elliptic & $(W,H,1)$ & $8$ \\
			Parabolic & $(W,H,1)$ & $8$ \\
			Coupled  & $(W,H,2)$ & $40$ \\ \hline
		\end{tabular}
	\end{minipage}\hfill
	\begin{minipage}{0.53\linewidth}
		\centering
		\includegraphics[width=0.7\linewidth]{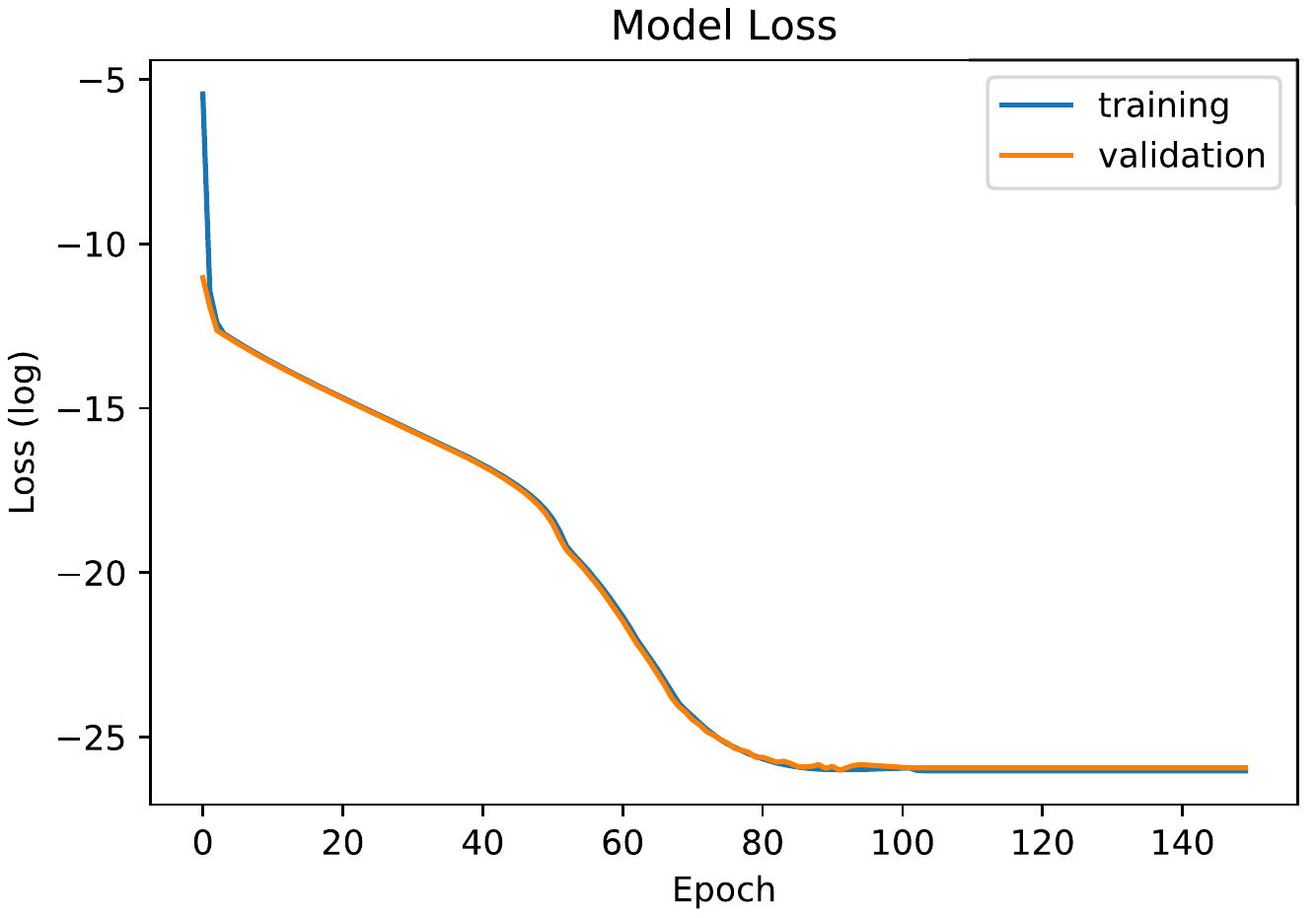}
		\captionof{figure}{MSE loss ($ln$) during training.}
		\label{fig:loss}
	\end{minipage}
\end{table}

\section{Kernel Representation and Interpretability} \label{sec:ker_rep}
	
\begin{figure}
	\centering
	\includegraphics[width=\linewidth]{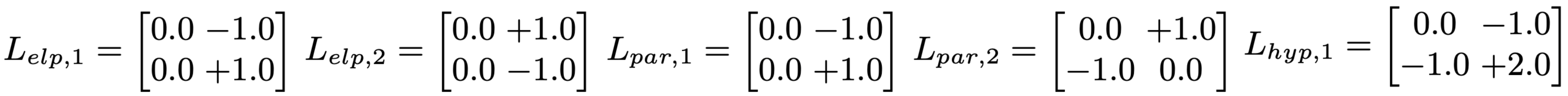}
	\caption{Isolated process kernels, rounded to first decimal. $L_{fam,lay}$ represents the learned CNN kernels for family $fam \in \mathcal{F}$ and $lay$ represents the depth. $\mathcal{F}=\{elp, par, hyp\}$ represents elliptic, parabolic, and hyperbolic families.}
	\label{fig:pure_ker}
\end{figure}

Let us now discuss kernel representations with emphasis on the choice of kernel size, number of kernels and network depth. We again consider the simple linear hyperbolic problem Eq. \ref{eqn:hyp} to identify difficulties and pitfalls in the design of a general architecture. These arguments apply equally to the kernels for isolated and coupled problems discussed before. Fig. \ref{fig:pure_ker} presents lowest order ($\mathcal{H}$ in $\mathcal{L_H}$) kernels ($L$) for the isolated processes used for verifying the learned kernels.

An immediate question arises from the choice of only one kernel for the linear hyperbolic problem at hand. To explain this point let us consider an equivalent choice of kernels for this problem as follows:
	\begin{align*}
	&L_{hyp}^{11} = \begin{bmatrix}
	0.0 & -1.0 \\ 
	0.0 & +1.0
	\end{bmatrix}
	L_{hyp}^{21} = \begin{bmatrix}
	0.0 & 0.0 \\ 
	+1.0 & -1.0
	\end{bmatrix}; 
	&L_{hyp}^{12} = \begin{bmatrix}
	0.0 & 0.0 \\ 
	0.0 & -1.0
	\end{bmatrix}
	L_{hyp}^{22} = \begin{bmatrix}
	0.0 & 0.0 \\ 
	0.0 & +1.0
	\end{bmatrix}
	\end{align*}
This equivalent representation is constructed by decomposing the original kernel $L_{hyp,1}$ into separate $\nabla_{x}$ ($L_{hyp,11}$) and $\nabla_{t}$ ($L_{hyp,11}$) parts with $L_{hyp,12}$ and $L_{hyp,12}$ serving as the construction step. In other words a $depth = 1$ convolutional network as well as a $depth = 2$ network described above are equally capable of representing the problem. The reason we select the single layer network over this two-layer equivalent network is two-fold: (1) The second representation makes the kernels sparse, and (2) our convolutional architecture relies upon the depth of the network to infer the order of the differential operator required to approximate the problem. 
	
The first point is clear from a simple comparison between the $depth = 2$ and $depth = 1$ representations. Consequently, converging to the true-kernels (global minimum) is computationally more intensive; as in VGG-Net \cite{simonyan2014very}, starting from an arbitrary initial guess for the network weights. The second argument allows us to easily interpret and explain our networks weights by relating it to an infinite dimensional PDE representation. This choice provides us further transparency where the overall representation $(\mathcal{L}_H)$ can then be validated by a human with the kernel weights leading to an approximate PDE expression.

Let us also consider a counter argument against the first point: Why not choose a regularization (kernel or overall) for the second representation? A valid argument however, notice that the choice of regularization requires a priori knowledge in the application domain. Moreover, the requirement is rather harsh \ie a regularization that will penalize deviation from zero for select kernel weights. Our architecture avoids these choices by favoring dense kernels in an effort to extract a low-weights and compact representation. In other words, we use dense kernels as opposed to explicit regularization.
	
\begin{figure}
		\centering
		\includegraphics[width=\linewidth]{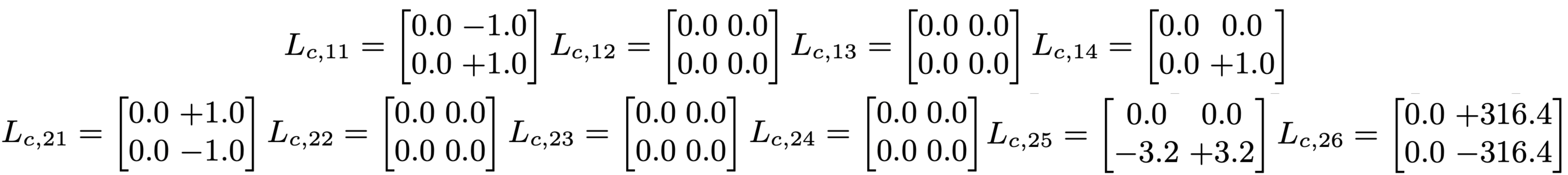}
		\caption{Coupled process kernels, rounded to first decimal.}
		\label{fig:ker_coupled}
	\end{figure}

As for the size of the kernel; $(2 \times 2)$ in our architecture, represents the lowest order approximation $\mathcal{H}$ in $\mathcal{L_H}$ of the differential operators (spatial or temporal). Again, the sparsity argument applies here as well by considering a second order approximation and the resulting $(3 \times 3)$ kernel for a $depth=1$ representation of the hyperbolic problem below:  
	\begin{align}
	\tilde{L}_{hyp} = \begin{bmatrix}
	0.0 & 0.0 & -1.0 \\ 
	0.0 & 0.0 & 0.0\\
	-1.0 &0.0 & 2.0
	\end{bmatrix}.
	\end{align}
In order to save space, we curtail equivalent higher depth representations of other isolated processes. Please note that our network design is only guided by the knowledge of these isolated process kernels towards finding the coupled process kernels as shown in Fig. \ref{fig:ker_coupled}. The kernel weights can alter during training and above description does not imply that they are fixed.
\section{Ablation Study}
\subsection{Transparency \vs network design}
To validate transparency, we explore two main design choices in our network: (1) effect of depth and (2) effect of \#kernels at a given depth. These two effects are mutually independent and we show that misrepresentation in each of them leads to loss of interpretation both in terms of the process and the accuracy of final results achieved. We look at the learned kernels, activation maps errors and training error under different network design. To set up the benchmark, let us consider the elliptic problem; Eq. \ref{eqn:elp} with kernels in Fig. \ref{fig:pure_ker} as $L_{elp,1}$ and $L_{elp,2}$ at $depth=1$ and $depth=2$. These size $(2,2)$ kernels compose to form a size $(3,3)$ elliptic kernel by simple valid convolution as follows:
	\begin{align}
	\tilde{L}_{elp} = \begin{bmatrix}
	0.0 & 0.0 & -1.0 \\ 
	0.0 & 0.0 & +2.0\\
	0.0 & 0.0 & -1.0
	\end{bmatrix} = 
	\begin{bmatrix}
	0.0 & -1.0 \\ 
	0.0 & +1.0
	\end{bmatrix} \bigotimes
	\begin{bmatrix}
	0.0 & +1.0 \\ 
	0.0 & -1.0
	\end{bmatrix}
	\end{align}
As discussed in Section \ref{sec:ker_rep}, searching for this sparse kernel is compute intensive and therefore we rely on the two decomposed $(2,2)$ kernels to reduce the sparsity. 
\begin{figure}[ht]
	\centering
	\begin{subfigure}[b]{0.15\linewidth}
		\centering
		$\begin{bmatrix}
		+0.2 & +0.7 \\
		-0.2 & -0.7
		\end{bmatrix}$
		\caption{Depth=1}
	\end{subfigure}
	\begin{subfigure}[b]{0.31\linewidth}
		\centering
		$\begin{bmatrix}
		0.0 & -1.0 \\
		0.0 & +1.0
		\end{bmatrix}$ 
		$\begin{bmatrix}
		0.0 & +1.0 \\
		0.0 & -1.0
		\end{bmatrix}$
		\caption{Depth=2}
	\end{subfigure}
	\begin{subfigure}[b]{0.51\linewidth}
		\centering
		$\begin{bmatrix}
		-0.4 & -0.1 \\
		+0.4 & +0.1
		\end{bmatrix}$
		$\begin{bmatrix}
		+0.1 & +0.2 \\
		-0.4 & +0.1
		\end{bmatrix}$
		$\begin{bmatrix}
		-0.5 & -0.2 \\
		+0.5 & +0.2
		\end{bmatrix}$
		\caption{Depth=3}
	\end{subfigure}
	\caption{Learned kernels for Elliptic PDE at different depths each having one kernel. Only the kernels learnt for depth 2 can represent the elliptic problem.}
	\label{fig:elp_ker_ablation}
\end{figure}
 
 \noindent \textbf{(1) Effect of depth:} We vary the network depth for a fixed order system to demonstrate that the change in performance is a function of the architecture shape and not attributed to an increase in capacity with a shallow/deeper network. To make this comparison, we train the elliptic network with depths one and three, each having one \textit{conv} filter at each layer. The learnt kernels are shown in Fig. \ref{fig:elp_ker_ablation}. The training error and activation errors are tabulated in Table \ref{tab:elp_error}. As observed from this table, a compact representation always performs better in compared to other design choices while maintaining interpretability.

\begin{table}[h]
	\centering
	\caption{Training details for elliptic pde for different network depths. The network of depth $2$ performs the best since it captures the second-order process}
	\begin{tabular}{|c|c|c|c|} \hline
		Depth & 1 & 2 & 3 \\ \hline
		Training Error & 1.8e-4 & 2.4e-11 & 5.6e-8 \\
		Activation Error & e-2 & e-7 & e-4 \\ 
		Epochs & 3 & 60 & 120 \\ \hline
	\end{tabular}
	\label{tab:elp_error}
\end{table}
	
\noindent \textbf{(2) Effect of \#kernels at a given depth:} To explore the effect of varying number of filters at a given layer on a fixed order system, we train the elliptic network. We hold the kernel at $depth=2$ and vary the number of kernels at $depth=1$. The learnt kernels are shown in Fig. \ref{fig:elp_ker_ablation_width}.
\begin{figure}[ht]
	\centering
	\begin{subfigure}[b]{0.4\linewidth}
		\centering
		$\begin{bmatrix}
		0.45 & -0.01 \\
		-0.24 & -0.19
		\end{bmatrix}$
		$\begin{bmatrix}
		0.10 & 0.26 \\
		0.26 & -0.63
		\end{bmatrix}$
		\caption{\#Kernels=2, Layer=1}
	\end{subfigure}
	\begin{subfigure}[b]{0.59\linewidth}
		\centering
		$\begin{bmatrix}
		0.50 & 0.1 \\
		0.21 & -0.81
		\end{bmatrix}$ 
		$\begin{bmatrix}
		-0.03 & -0.76 \\
		0.25 & 0.54
		\end{bmatrix}$
		$\begin{bmatrix}
		-0.12 & -0.23 \\
		0.56 & -0.21
		\end{bmatrix}$
		\caption{\#Kernels=3, Layer=1}
	\end{subfigure}
	\caption{Learned kernels for Elliptic PDE for multiple kernels at $depth=1$. Both cases result in training loss of $\sim e-7, \sim e-6$, respectively showing that they were unable to capture the true elliptic kernels at the first layer.}
	\label{fig:elp_ker_ablation_width}
\end{figure}

\subsection{Network explanation against scaling and missing data}
\begin{figure}
    \centering
		\includegraphics[width=0.9\linewidth]{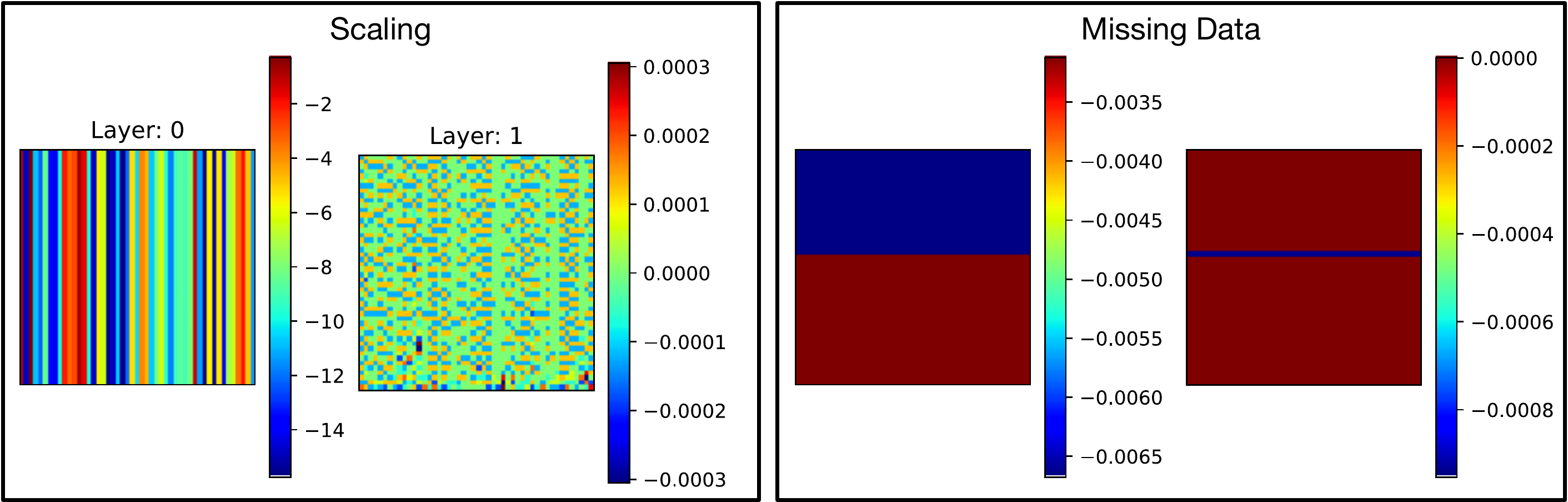}
    \caption{Effect of scaling (left) and missing data (right) in the activation maps.}
    \label{fig:act_explain}
\end{figure}

Here, we perform a sensitivity analysis of the network for the following two cases:

\noindent \textbf{(1) Effect of scaling:} We tested the network with an image from the elliptic problem. The image was scaled by $1000$ to study the effect of scaling in the activation maps. The scaling translates linearly in the maps and can be seen easily by comparing Fig. \ref{fig:act_explain} (left) and the elliptic activation maps in Fig. \ref{fig:act_isolated}.

\noindent \textbf{(2) Effect of missing data:} We take the parabolic network and test it on an image with missing data in the form of external perturbation. The perturbation is introduced  spatially in the middle for all times. The network identifies this perturbation or missing information as a horizontal feature in the activation maps as deviations from the learned features. Fig. \ref{fig:act_explain} (right) shows the spatial and temporal span of the perturbation.
	
\section{Conclusion}

We demonstrate the effectiveness of our low-weight, adaptive network for various isolated/coupled system of PDEs. Any modeler can easily verify the network workings by ratifying their belief (assumed form of PDE) against the \#kernels and network depth, to validate transparency and build user trust. Interpretability is achieved by composing the learned kernels for a discrete representation of the PDE. The network is explainable since the effects of scaling and missing data visually manifest in the activation maps during testing. Our framework combines these three characteristics for a robust evolution of current understanding from isolated to coupled processes. The outlined design strategy is flexible and will imbibe complicated processes with additional domain constraints in future works.

\bibliographystyle{unsrt}
\bibliography{references}

\end{document}